# Approximating Higher-Order Distances Using Random Projections

**Ping Li**[*]
Department of Statistical Science
Faculty of Computing and Information Science
Cornell University, Ithaca, NY 14853
pingli@cornell.edu

**Michael W. Mahoney**[†]
Department of Mathematics
Stanford University
Stanford, CA 94305
mmahoney@cs.stanford.edu

**Yiyuan She**
Department of Statistics
Florida State University
Tallahassee, FL 32306
yshe@stat.fsu.edu

## Abstract

We provide a simple method and relevant theoretical analysis for efficiently estimating higher-order $l_p$ distances. While the analysis mainly focuses on $l_4$, our methodology extends naturally to $p = 6, 8, 10...$, (i.e., when $p$ is even).

Distance-based methods are popular in machine learning. In large-scale applications, storing, computing, and retrieving the distances can be both space and time prohibitive. Efficient algorithms exist for estimating $l_p$ distances if $0 < p \leq 2$. The task for $p > 2$ is known to be difficult. Our work partially fills this gap.

## 1 Introduction

Distance-based methods are popular in machine learning, for example, nearest neighbor methods, kernel SVM, multidimensional scaling, etc. It is often the case that choosing an appropriate distance may be critical to the performance. This study concerns the $l_p$ distance. Given two data vectors $x, y \in \mathbb{R}^D$, we define their $l_p$ distance to be

$$d_{(p)} = \sum_{i=1}^{D} |x_i - y_i|^p. \quad (1)$$

While the Euclidian distance (i.e., $p = 2$) is the most popular, the use of the $l_1$ and $l_\infty$ distances is not uncommon. It is believed that using the $l_1$ distance may often produce more "robust" results. See [10] for an example of using the $l_\infty$ distance for object searching in high dimensions.

Distanced-based outlier/anomaly detection is popular [16]. When the data are normal, the distribution is determined by the first two moments. To identify interesting (or anomalous) components, it is often necessary to use higher-order moments; for example, the use of the $l_4$ distance in independent component analysis (ICA) [11]. The $l_4$ distance, closely related to the kurtosis, is an important summary statistic. In this spirit, using higher-order distances provides a tunable mechanism for effective anomaly detection.

[*]Supported by NSF-DMS, ONR-YIP, Microsoft, and Google.
[†]Supported by AFOSR.

### 1.1 Impact of $p$ on the Nearest Neighbor Method

The choice of $p$ can have a noticeable impact on the performance of learning algorithms [5]. To demonstrate this, an experiment of $m$-nearest neighbor classification was conducted on several datasets (see Table 1). Figure 1 reports the test classification errors for $p$ from 0.5 to 10.

Table 1: Datasets. For *Mnist10k*, we swapped the original *Mnist* training set with the test set. For *Realsim*, we took a 10% subset from the original data, which was then randomly split into a training set and a test set. For (UCI) *Gisette*, we used the original validation set as the test set because the test labels are missing.

| Dataset | # class | # training | # test | $D$ |
|---|---|---|---|---|
| Mnist | 10 | 60000 | 10000 | 784 |
| Letter | 26 | 15000 | 5000 | 16 |
| Mnist10k | 10 | 10000 | 60000 | 784 |
| Zipcode | 10 | 7291 | 2007 | 256 |
| RealSim | 2 | 6000 | 1128 | 20958 |
| Gisette | 2 | 6000 | 1000 | 5000 |

Figure 1 demonstrates that using the appropriate $l_p$ distance may have substantial impact on the classification performance. The "optimal" $p$ depends on the data (and $m$). For *Letter*, the $l_2$ distance is the best choice. For other datasets, however, the smallest test errors usually occur when $p \geq 4$.

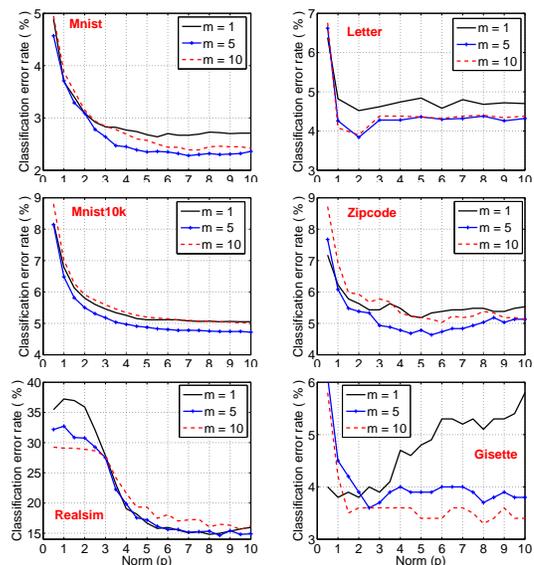

Figure 1: Test error rates using $m$-nearest neighbors, to demonstrate the significant impact of the norm ($p$).

## 1.2 Computational & Storage Burden

Learning algorithms often assume a data matrix $\mathbf{A} \in \mathbb{R}^{n \times D}$ ($n$ observations and $D$ dimensions). For distance-based methods, a basic task is to compute, store, and retrieve the distances [25], which becomes non-trivial when the data matrix (both $n$ and $D$) is large. For kernel SVM, storing and computing kernels is often the bottleneck [4].

For example, if $\mathbf{A}$ is a Web term-doc matrix with each row representing one Web page, then $n > 10^{10}$ and it is possible that $D = 10^6$ (single words) or $D = 10^{18}$ (3-shingles). Another example is the image data. Using pixels as features, a $1000 \times 1000$ image can be represented by a vector of dimension $D = 10^6$. Using histogram-based features, e.g., [5], $D = 256^3 = 16,777,216$ is possible if one discretizes the RGB space into $256^3$ scales.

In large-scale applications, the data matrix may be too large for the physical memory. Often it may be also infeasible to store the distance (similarity) matrix at the cost of $O(n^2)$. For example, when $n = 10^6$ (which is not the largest application in these days), the memory cost of $O(10^{12})$ can be infeasible. A commonly used strategy [4] is to compute the distances *on demand*. This strategy, however, is only desirable when (A): the data matrix can be stored in the memory, and (B): the additional computational overhead of the *on-demand* strategy should be small enough.

## 1.3 Random Projections for Estimating $l_2$ Distances

The method of (normal) random projections [26] has become a standard technique for efficiently computing the $l_2$ distances in machine learning [3, 8, 21, 6, 9]. The idea is to multiply the original data matrix $\mathbf{A} \in \mathbb{R}^{n \times D}$ by a random matrix $\mathbf{R} \in \mathbb{R}^{D \times k}$, resulting in a much smaller matrix $\mathbf{B} = \mathbf{A} \times \mathbf{R} \in \mathbb{R}^{n \times k}$. Entries of $\mathbf{R}$ are sampled i.i.d. from the standard normal: $r_{ij} \sim N(0, 1)$. The classical *Johnson-Lindenstrauss (JL)* Lemma [14, 7] says it suffices to let $k = O\left(\log n / \epsilon^2\right)$ so that all pairwise $l_2$ distances can be estimated within a $1 \pm \epsilon$ factor with high probability.

Similar to the $l_2$ case, the method of *stable random projections* [12, 18, 17] is an effective algorithm for computing the $l_p$ distances for $0 < p \leq 2$, by sampling $\mathbf{R}$ from a $p$-stable distribution. The required number of projections is still $O\left(\log n / \epsilon^2\right)$, which is again independent of $D$.

## 1.4 Dimension Reduction in $l_p$ with $p > 2$

As proved in the Theory literature [2, 24], for $p > 2$, the (worst case) sample size lower bound is $\Omega\left(D^{1-2/p}\right)$, which increases with increasing $D$, unlike in the case of $0 < p \leq 2$.

The lower bound is merely a theoretical limit. Developing algorithms (especially practical algorithms) to achieve the bound is a separate issue. When there is only one vector (i.e., $n = 1$) or a pair of vectors ($n = 2$) in $D$ dimensions, [13] proposed an algorithm to approach the theoretical limit, in the context of data streams.

## 1.5 Our Proposed Approach

Our idea is simple, based on the fact that, when $p$ is even, the $l_p$ distance can be decomposed into marginal $l_p$ norms and various "inner products." For example, when $p = 4$,

$$d_{(4)} = \sum_{i=1}^D x_i^4 + \sum_{i=1}^D y_i^4 + 6\sum_{i=1}^D x_i^2 y_i^2 - 4\sum_{i=1}^D x_i^3 y_i - 4\sum_{i=1}^D x_i y_i^3.$$

We assume a linear scan of the data matrix is feasible. As an option, the marginal norms, $\sum_{i=1}^D x_i^4$ and $\sum_{i=1}^D y_i^4$, can be computed exactly by one scan. We propose using (normal) random projections to estimate these inner products: $\sum_{i=1}^D x_i^2 y_i^2$, $\sum_{i=1}^D x_i^3 y_i$, $\sum_{i=1}^D x_i y_i^3$. We will have the choice of either using 3 independent random projection matrices to estimate the three inner products separately, or using only 1 random projection matrix to estimate all 3. We will provide theoretical analysis for both methods.

Our method extends naturally to $p = 6, 8, 10, \ldots$ Note that, since the sample size is lower bounded by $\Omega\left(D^{1-2/p}\right)$, larger $p$ requires significantly more samples.

This method has the meritable simplicity. Retrospectively, it is also a bit surprising that dimension reduction in $l_p$ with $p > 2$ can be partially solved using the method of normal random projections, which was meant only for $l_2$.

## 2 Two Baselines Algorithms Using Sampling

### 2.1 Simple Random Sampling

We can randomly sample $k$ columns from the data matrix $\mathbf{A} \in \mathbb{R}^{n \times D}$, to compute the $l_4$ distances. Consider, for example, two rows in $\mathbf{A}$, denoted by $x$ and $y$. Denote the sampled $k$ entries by $\tilde{x}_j$, $\tilde{y}_j$, $j = 1$ to $k$. The estimator, denoted by $\hat{d}_{(4),S}$, is

$$\hat{d}_{(4),S} = \frac{D}{k} \sum_{j=1}^k |\tilde{x}_j - \tilde{y}_j|^4 \quad (2)$$

$$\text{Var}\left(\hat{d}_{(4),S}\right) = \frac{D^2}{k^2} k \left[E\left(|\tilde{x}_j - \tilde{y}_j|^8\right) - E^2\left(|\tilde{x}_j - \tilde{y}_j|^4\right)\right]$$

$$= \frac{D}{k} \left(\sum_{i=1}^D |x_i - y_i|^8 - \frac{\left(\sum_{i=1}^D |x_i - y_i|^4\right)^2}{D}\right). \quad (3)$$

In the worst case, the variance is dominated by the 8th order terms. This is why random sampling often has very large errors, although it is the widely-used default method.

### 2.2 Conditional Random Sampling (CRS)

Conditional Random Sampling (CRS) [19, 20] was recently proposed for sampling from sparse data. Due to the space limit, we will not elaborate on this algorithm. Denote the CRS estimate by $\hat{d}_{(4),CRS}$. The variance is

$$\text{Var}\left(\hat{d}_{(4),CRS}\right) \approx \frac{\max\{|x|_0, |y|_0\}}{D} \times \text{Var}\left(\hat{d}_{(4),S}\right), \quad (4)$$

where $|x|_0$ and $|y|_0$ are the numbers of non-zeros in vectors $x$ and $y$, respectively. While the variance of CRS is reduced substantially by taking advantage of the data sparsity, it is nevertheless still dominated by the 8th order terms.

# 3 The Proposed Algorithms

This section presents our proposed algorithms based on random projections for estimating the $l_4$ distance.

## 3.1 Using 3 Normal Random Projections Matrices

We independently generate three normal random matrices of size $D \times k$: $\mathbf{R}^{(1)}$, $\mathbf{R}^{(2)}$, $\mathbf{R}^{(3)}$; entries are, respectively denoted by $r_{ij}^{(1)}$, $r_{ij}^{(2)}$, $r_{ij}^{(3)}$, $i = 1$ to $D$, $j = 1$ to $k$. For two vectors $x, y \in \mathbb{R}^D$, we generate $u_1, u_2, u_3, v_1, v_2, v_3 \in \mathbb{R}^k$:

$$u_{1,j} = \sum_{i=1}^{D} x_i r_{ij}^{(1)}, \quad u_{2,j} = \sum_{i=1}^{D} x_i^2 r_{ij}^{(2)}, \quad u_{3,j} = \sum_{i=1}^{D} x_i^3 r_{ij}^{(3)},$$

$$v_{1,j} = \sum_{i=1}^{D} y_i r_{ij}^{(1)}, \quad v_{2,j} = \sum_{i=1}^{D} y_i^2 r_{ij}^{(2)}, \quad v_{3,j} = \sum_{i=1}^{D} y_i^3 r_{ij}^{(3)}.$$

We have an unbiased estimator, denoted by $\hat{d}_{(4),3p}$

$$\hat{d}_{(4),3p} = \sum_{i=1}^{D} x_i^4 + \sum_{i=1}^{D} y_i^4 + \frac{1}{k}(6u_2^T v_2 - 4u_3^T v_1 - 4u_1^T v_3) \quad (5)$$

**Lemma 1**

$$E\left(\hat{d}_{(4),3p}\right) = d_{(4)},$$

$$\text{Var}\left(\hat{d}_{(4),3p}\right) = \frac{36}{k}\left(\sum_{i=1}^{D} x_i^4 \sum_{i=1}^{D} y_i^4 + \left(\sum_{i=1}^{D} x_i^2 y_i^2\right)^2\right) \quad (6)$$

$$+ \frac{16}{k}\left(\sum_{i=1}^{D} x_i^6 \sum_{i=1}^{D} y_i^2 + \left(\sum_{i=1}^{D} x_i^3 y_i\right)^2\right)$$

$$+ \frac{16}{k}\left(\sum_{i=1}^{D} x_i^2 \sum_{i=1}^{D} y_i^6 + \left(\sum_{i=1}^{D} x_i y_i^3\right)^2\right).$$

*Proof:* See the proof of Lemma 6 in Appendix A. □

## 3.2 Using 3 Projection Matrices and Margins

Since we assume that a linear scan of the data is feasible, we can (optionally) take advantage of the marginal norms, $\sum_{i=1}^{D} x_i^2$, $\sum_{i=1}^{D} x_i^4$, $\sum_{i=1}^{D} x_i^6$, etc, to reduce the variance. Lemma 2 demonstrates such an estimator.

**Lemma 2** We can estimate $d_{(4)}$ by $\hat{d}_{(4),3p,m}$, where

$$\hat{d}_{(4),3p,m} = \sum_{i=1}^{D} x_i^4 + \sum_{i=1}^{D} y_i^4 + 6\hat{a}_{2,2} - 4\hat{a}_{3,1} - 4\hat{a}_{1,3}, \quad (7)$$

and $\hat{a}_{2,2}$, $\hat{a}_{3,1}$, $\hat{a}_{1,3}$, are respectively, the solutions to the following three cubic equations:

$$a_{2,2}^3 - \frac{a_{2,2}^2}{k} u_2^T v_2 - \frac{1}{k} \sum_{i=1}^{D} x_i^4 \sum_{i=1}^{D} y_i^4 u_2^T v_2 - a_{2,2} \left(\sum_{i=1}^{D} x_i^4 \sum_{i=1}^{D} y_i^4\right)$$

$$+ \frac{a_{2,2}}{k}\left(\sum_{i=1}^{D} x_i^4 \|v_2\|^2 + \sum_{i=1}^{D} y_i^4 \|u_2\|^2\right) = 0.$$

$$a_{3,1}^3 - \frac{a_{3,1}^2}{k} u_3^T v_1 - \frac{1}{k} \sum_{i=1}^{D} x_i^6 \sum_{i=1}^{D} y_i^2 u_3^T v_1 - a_{3,1}\left(\sum_{i=1}^{D} x_i^6 \sum_{i=1}^{D} y_i^2\right)$$

$$+ \frac{a_{3,1}}{k}\left(\sum_{i=1}^{D} x_i^6 \|v_1\|^2 + \sum_{i=1}^{D} y_i^2 \|u_3\|^2\right) = 0.$$

$$a_{1,3}^3 - \frac{a_{1,3}^2}{k} u_1^T v_3 - \frac{1}{k} \sum_{i=1}^{D} x_i^2 \sum_{i=1}^{D} y_i^6 u_1^T v_3 - a_{1,3}\left(\sum_{i=1}^{D} x_i^2 \sum_{i=1}^{D} y_i^6\right)$$

$$+ \frac{a_{1,3}}{k}\left(\sum_{i=1}^{D} x_i^2 \|v_3\|^2 + \sum_{i=1}^{D} y_i^6 \|u_1\|^2\right) = 0.$$

*Asymptotically (as $k \to \infty$), the variance would be*

$$\text{Var}\left(\hat{d}_{(4),3p,m}\right)$$

$$= 36\text{Var}(\hat{a}_{2,2}) + 16\text{Var}(\hat{a}_{2,2}) + 16\text{Var}(\hat{a}_{2,2}) \quad (8)$$

$$= \frac{36}{k} \frac{\left(\sum_{i=1}^{D} x_i^4 \sum_{i=1}^{D} y_i^4 - \left(\sum_{i=1}^{D} x_i^2 y_i^2\right)^2\right)^2}{\sum_{i=1}^{D} x_i^4 \sum_{i=1}^{D} y_i^4 + \left(\sum_{i=1}^{D} x_i^2 y_i^2\right)^2}$$

$$+ \frac{16}{k} \frac{\left(\sum_{i=1}^{D} x_i^6 \sum_{i=1}^{D} y_i^2 - \left(\sum_{i=1}^{D} x_i^3 y_i\right)^2\right)^2}{\sum_{i=1}^{D} x_i^6 \sum_{i=1}^{D} y_i^2 + \left(\sum_{i=1}^{D} x_i^3 y_i\right)^2}$$

$$+ \frac{16}{k} \frac{\left(\sum_{i=1}^{D} x_i^2 \sum_{i=1}^{D} y_i^6 - \left(\sum_{i=1}^{D} x_i y_i^3\right)^2\right)^2}{\sum_{i=1}^{D} x_i^2 \sum_{i=1}^{D} y_i^6 + \left(\sum_{i=1}^{D} x_i y_i^3\right)^2} + O\left(\frac{1}{k^2}\right)$$

*Proof:* [21] proposed using the marginal $l_2$ norms to improve the estimates of $l_2$ distances and inner products. Because our three projection matrices are independent, we can use a similar procedure to analyze $\hat{a}_{2,2}$, $\hat{a}_{3,1}$, and $\hat{a}_{1,3}$, independently and then combine the results.

## 3.3 Using Only 1 Random Projection Matrix

Using 3 random projection matrices simplifies the theoretical analysis. However, it would be practically much more convenient if we just need to use one random matrix.

Here, we generate only one random matrix of $\mathbf{R} \in \mathbb{R}^{D \times k}$ with i.i.d. entries $r_{ij} \sim N(0,1)$, and six vectors $\in \mathbb{R}^k$:

$$u_{1,j} = \sum_{i=1}^{D} x_i r_{ij}, \quad u_{2,j} = \sum_{i=1}^{D} x_i^2 r_{ij}, \quad u_{3,j} = \sum_{i=1}^{D} x_i^3 r_{ij},$$

$$v_{1,j} = \sum_{i=1}^{D} y_i r_{ij}, \quad v_{2,j} = \sum_{i=1}^{D} y_i^2 r_{ij}, \quad v_{3,j} = \sum_{i=1}^{D} y_i^3 r_{ij}.$$

We again have a simple unbiased estimator of $d_{(4)}$

$$\hat{d}_{(4),1p} = \sum_{i=1}^{D} x_i^4 + \sum_{i=1}^{D} y_i^4 + \frac{1}{k}(6u_2^T v_2 - 4u_3^T v_1 - 4u_1^T v_3) \quad (9)$$

**Lemma 3**

$$E\left(\hat{d}_{(4),1p}\right) = d_{(4)},$$

$$\text{Var}\left(\hat{d}_{(4),1p}\right) = \text{Var}\left(\hat{d}_{(4),3p}\right) + \Delta_{1p} \quad (10)$$

$$\Delta_{1p} = -\frac{48}{k}\left(\sum_{i=1}^{D} x_i^5 \sum_{i=1}^{D} y_i^3 + \sum_{i=1}^{D} x_i^2 y_i \sum_{i=1}^{D} x_i^3 y_i^2\right) \quad (11)$$

$$- \frac{48}{k}\left(\sum_{i=1}^{D} x_i^3 \sum_{i=1}^{D} y_i^5 + \sum_{i=1}^{D} x_i y_i^2 \sum_{i=1}^{D} x_i^2 y_i^3\right)$$

$$+ \frac{32}{k}\left(\sum_{i=1}^{D} x_i^4 \sum_{i=1}^{D} y_i^4 + \sum_{i=1}^{D} x_i y_i \sum_{i=1}^{D} x_i^3 y_i^3\right).$$

*Proof:* See the proof of Lemma 6 in Appendix A. □

## 3.4 Using 1 Projection Matrix and Margins

We can also take advantage of the marginal norms just like in Sec 3.2. The resultant estimator, denoted by $\hat{d}_{(4),1p,m}$, has exactly the same formula as $\hat{d}_{(4),3p,m}$. The variance of $\hat{d}_{(4),1p,m}$ is not as easy to analyze due to the correlations.

Note that $\hat{d}_{(4),1p,m}$ is not the full MLE solution. If we write down the covariance matrix, it is no longer block diagonal. This makes it very difficult to find the full MLE solution.

### 3.5 1 Projection Matrix Can Be Better than 3

Using only 1 projection matrix is much simpler than using 3 projection matrices. Moreover, we can show that, when the data are nonnegative (which is often the case in the real-world), it is often better (although not always) to use only 1 projection matrix, according to the following Lemma.

**Lemma 4** *Suppose the data are nonnegative: $x_i \geq 0, y_i \geq 0$. We always have $\text{Var}\left(\hat{d}_{(4),3p}\right) \geq \text{Var}\left(\hat{d}_{(4),1p}\right)$, i.e., $\Delta_{1p} \leq 0$ in (11), as long as the following relation holds:*

$$5\sum_{i=1}^{D} z_i^3 \sum_{i=1}^{D} z_i^5 - \sum_{i=1}^{D} z_i^2 \sum_{i=1}^{D} z_i^6 \geq 0, \quad z_i = \sqrt{x_i y_i}. \quad (12)$$

***Proof:*** *The proof is omitted due to the space limit.* □

While in theory the condition (12) needs not hold for all $z_i \geq 0$, the inequality is often true in "reasonable" data. Since $D$ is large, we can essentially write (12) in an expectation form

$$5E(z^3)E(z^5) - E(z^2)E(z^6) \geq 0. \quad (13)$$

**Lemma 5** *The inequality (13) always holds if $z$ follows a Gamma distribution or a Beta distribution.*

***Proof:*** *The proof is omitted due to the space limit.* □

The Gamma distribution has a semi-infinite support and includes the exponential and Chi-square distributions as special cases. The Beta distribution has a finite support and includes the uniform distribution as a special case.

### 3.6 Improving Estimates for Nearly Identical Vectors

We have presented several estimators using random projections: $\hat{d}_{(4),3p}$, $\hat{d}_{(4),3p,m}$, $\hat{d}_{(4),1p}$, and $\hat{d}_{(4),1p,m}$, and derived the variances for the first three estimators, i.e., (6), (8), and (10). However, when the two original vectors, $x$ and $y$, are nearly identical (i.e., $x_i \approx y_i$), these variances become small but do not approach zero. Although in practice we do not expect this will be a major concern, it is nevertheless an interesting theoretical task to develop estimators whose variances approach zero when $x_i \approx y_i$. We have found such an estimator:

$$\hat{d}_{(4),1p,I} = \frac{1}{k}\left(-3u_2^T u_2 - 3v_2^T v_2 + 4u_1^T u_3 + 4v_1^T v_3\right)$$
$$+ \frac{1}{k}(6u_2^T v_2 - 4u_3^T v_1 - 4u_1^T v_3) \quad (14)$$

Compared with $\hat{d}_{(4),1p}$, here we basically replace the exact computation $\sum_{i=1}^{D} x_i^4 + \sum_{i=1}^{D} y_i^4$ by the approximation $\frac{1}{k}\left(-3u_2^T u_2 - 3v_2^T v_2 + 4u_1^T u_3 + 4v_1^T v_3\right)$. This may initially appear strange, but a careful examination reveals such an estimator is natural.

When $x_i = y_i$ exactly, then $u_1 = v_1$, $u_2 = v_2$, and $u_3 = v_3$. In this case, $\hat{d}_{(4),1p,I} = 0$, as desired. This estimator $\hat{d}_{(4),1p,I}$ is a nice example of utilizing error cancelations.

**Lemma 6**

$$E\left(\hat{d}_{(4),1p,I}\right) = d_{(4)},$$

$$\text{Var}\left(\hat{d}_{(4),1p,I}\right) = \text{Var}\left(\hat{d}_{(4),1p}\right) + \Delta_I, \quad (15)$$

$$\Delta_I = \frac{36}{k}\left(\sum_{i=1}^{D} x_i^2 y_i^2\right)^2 + \frac{34}{k}\left(\left(\sum_{i=1}^{D} x_i^4\right)^2 + \left(\sum_{i=1}^{D} y_i^4\right)^2\right) \quad (16)$$

$$+ \frac{32}{k}\sum_{i=1}^{D} x_i y_i^3 \sum_{i=1}^{D} x_i^3 y_i - \frac{32}{k}\sum_{i=1}^{D} x_i^4 \sum_{i=1}^{D} x_i^3 y_i$$

$$- \frac{72}{k}\left(\sum_{i=1}^{D} x_i^4 \sum_{i=1}^{D} x_i^2 y_i^2 + \sum_{i=1}^{D} y_i^4 \sum_{i=1}^{D} x_i^2 y_i^2\right)$$

$$- \frac{32}{k}\left(\sum_{i=1}^{D} y_i^4 \sum_{i=1}^{D} x_i^3 y_i + \sum_{i=1}^{D} x_i^4 \sum_{i=1}^{D} x_i y_i^3 + \sum_{i=1}^{D} y_i^4 \sum_{i=1}^{D} x_i y_i^3\right)$$

$$- \frac{48}{k}\left(\sum_{i=1}^{D} x_i^3 \sum_{i=1}^{D} x_i^5 + \sum_{i=1}^{D} x_i^2 y_i \sum_{i=1}^{D} x_i^2 y_i^3\right)$$

$$- \frac{48}{k}\left(\sum_{i=1}^{D} y_i^3 \sum_{i=1}^{D} y_i^5 + \sum_{i=1}^{D} x_i y_i^2 \sum_{i=1}^{D} x_i^3 y_i^2\right)$$

$$+ \frac{48}{k}\left(\sum_{i=1}^{D} x_i^3 \sum_{i=1}^{D} x_i^3 y_i^2 + \sum_{i=1}^{D} x_i^5 \sum_{i=1}^{D} x_i y_i^2 + \sum_{i=1}^{D} y_i^3 \sum_{i=1}^{D} x_i^2 y_i^3\right)$$

$$+ \frac{48}{k}\left(\sum_{i=1}^{D} y_i^5 \sum_{i=1}^{D} x_i^2 y_i + \sum_{i=1}^{D} x_i^5 \sum_{i=1}^{D} x_i^2 y_i + \sum_{i=1}^{D} y_i^3 \sum_{i=1}^{D} x_i^3 y_i^2\right)$$

$$+ \frac{48}{k}\left(\sum_{i=1}^{D} x_i^3 \sum_{i=1}^{D} x_i^2 y_i^3 + \sum_{i=1}^{D} y_i^5 \sum_{i=1}^{D} x_i y_i^2\right)$$

$$- \frac{32}{k}\left(\sum_{i=1}^{D} x_i^6 \sum_{i=1}^{D} x_i y_i + \sum_{i=1}^{D} y_i^2 \sum_{i=1}^{D} x_i^3 y_i^3\right)$$

$$- \frac{32}{k}\left(\sum_{i=1}^{D} x_i^2 \sum_{i=1}^{D} x_i^3 y_i^3 + \sum_{i=1}^{D} y_i^6 \sum_{i=1}^{D} x_i y_i\right)$$

$$+ \frac{16}{k}\left(\sum_{i=1}^{D} x_i^2 \sum_{i=1}^{D} x_i^6 + \sum_{i=1}^{D} y_i^2 \sum_{i=1}^{D} y_i^6 + 2\sum_{i=1}^{D} x_i y_i \sum_{i=1}^{D} x_i^3 y_i^3\right)$$

***Proof:*** *See Appendix A.* □

Clearly, when $x_i = y_i$, we always have $\text{Var}\left(\hat{d}_{1p,I}\right) = 0$, as expected. However, in general, it is not always the case that $\Delta_I \leq 0$. Thus, to use this estimator in practice, we have to combine it with $\hat{d}_{(4),1p}$ according to the observed $\Delta_I$. This condition is quite complicated.

## 4 Experiments

**Experiment 1** is a sanity check to validate the theoretical error (variance) predictions of the three estimation methods for the $l_4$ distances: simple random sampling, random projections using 3 projection matrices, and random projections using only 1 projection matrix.

In **Experiment 2** and **Experiment 3**, we ran the $m$-nearest neighbor algorithm for classifications using the estimated $l_4$ distances on the *Gisette* dataset and the *Realism* dataset.

### 4.1 Experiment 1

The Web crawl dataset consists of 10 pairs of word vectors. A binary-quantized version of the same dataset was used in [23]. For each word vector, the $j$th element is the number of times the word appeared in the $j$th Web page.

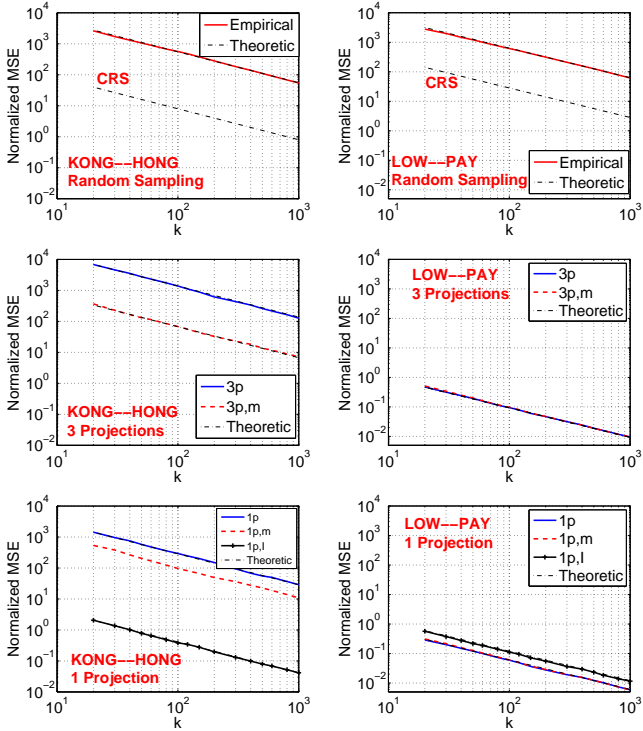

Figure 2: **Experiment 1**. MSE (Var+bias$^2$) normalized by the true squared $l_4$ distances for two word pairs: *KONG–HONG* (left panels) and *LOW–PAY* (right panels). The curves for the theoretical variances are essentially invisible (expect for CRS) because they overlap the empirical MSEs. **Top two panels (sampling methods):** There are three curves in each panel, including a solid curve for the empirical MSE of simple random sampling using $\hat{d}_{(4),S}$ (2) and two dot dashed curves for the theoretical variance (3) of simple random sampling and the theoretical variance (4) of CRS. Note that we did not conduct simulations for CRS. **Middle two panels (3 projection matrices):** There are four curves in each panel, although only two (or even just one, for *LOW–PAY*) curves are visible. The solid curve and dashed curve present the empirical MSEs using $\hat{d}_{(4),3p}$ (5) and $\hat{d}_{(4),3p,m}$ (7), respectively. The two (invisible) dot dashed curves are the corresponding theoretical variances (6) and (8) which overlap their empirical counterparts.
**Bottom two panels (1 projection matrix)**: Each panel contains five curves although at most three are visible. The plain solid curve and the solid curve marked by "+" present the empirical MSEs of the estimators $\hat{d}_{(4),1p}$ (9) and $\hat{d}_{(4),1p,I}$ (14), respectively. The dashed curve presents the empirical MSE of the estimator $\hat{d}_{(4),1p,m}$ using margins.

Table 2 summarizes the 10 word pairs, which include frequent word pairs (e.g., *OF–AND*), rare word pairs (e.g., *GAMBIA–KIRIBATI*), unbalanced pairs (e.g., *A–Test*), highly similar pairs (e.g, *KONG–HONG*), as well as word pairs that are not similar (e.g., *LOW–PAY*).

Table 2: Ten pairs of word vectors used in Experiment 1.

| Word 1 | Word 2 | Sparsity 1 | Sparsity 2 | $\beta_4$ | $\rho$ |
|---|---|---|---|---|---|
| KONG | HONG | 0.0145 | 0.0143 | 0.9832 | 0.9621 |
| OF | AND | 0.5697 | 0.5537 | 0.9314 | 0.8692 |
| SAN | FRANCISCO | 0.0487 | 0.0252 | 0.3547 | 0.5586 |
| UNITED | STATES | 0.0622 | 0.0607 | 0.3490 | 0.6653 |
| GAMBIA | KIRIBATI | 0.0031 | 0.0028 | 0.1645 | 0.5241 |
| CREDIT | CARD | 0.0458 | 0.0412 | 0.1453 | 0.2435 |
| RIGHTS | RESERVED | 0.1867 | 0.1720 | 0.0322 | 0.3480 |
| A | TEST | 0.5961 | 0.0348 | 0.0149 | 0.1296 |
| TIME | JOB | 0.1890 | 0.0498 | 0.0045 | 0.1530 |
| LOW | PAY | 0.0448 | 0.0432 | 0.0004 | 0.0460 |

In Table 2, sparsity is the fraction of non-zeros in each vector. The table also includes two measures of vector "similarity": the usual correlation coefficient $\rho$, which is suitable for $l_2$, and another measure designed for $l_4$:

$$\beta_4 = 1 - \frac{\sum_{i=1}^{D} |x_i - y_i|^4}{\sum_{i=1}^{D} x_i^4 + \sum_{i=1}^{D} y_i^4} \quad (17)$$

Figures 2 to 6 present the empirical mean square errors (MSE = var+bia$^2$, normalized by $d_{(4)}^2$.) for estimating the $l_4$ distances using various estimators we have presented:

- The theoretical variances, (3), (6), (8), (10), (15), are accurate, because they overlap the empirical MSEs.

- Simple random sampling usually has very large errors. Conditional Random Sampling (CRS) considerably reduces the variances by taking advantage of the data sparsity. Nevertheless, in most cases CRS still exhibits much larger errors than random projections.

- Using 1 projection matrix produces noticeably better estimates than using 3 projection matrices.

- Margin helps, especially for pairs of similar vectors.

- When two vectors are highly similar, using the special estimator $\hat{d}_{(4),1p,I}$ (14) produces substantially better estimates than all other estimators. However, when the two vectors are not too similar, using $\hat{d}_{(4),1p,I}$ may actually result in worse estimates than $\hat{d}_{(4),1p}$.

From these experiments, it should be clear that there is no need to use three projection matrices.

### 4.2 Experiment 2

In this experiment, we applied $m$-nearest neighbor method to classify the *Gisettee* dataset, using the estimated $l_4$ distances. Only the methods with 1 projection matrix, i.e., $\hat{d}_{(4),1p}$ and $\hat{d}_{(4),1p,m}$, were used. Figure 7 reports the experiment results:

- With $k > 500$ projections, using the estimated $l_4$ distances produces similar classification results as using the original $l_4$ distances.

- Using margins results in better classification results, with smaller average errors (i.e., more accurate) and smaller standard deviations (i.e., more stable).

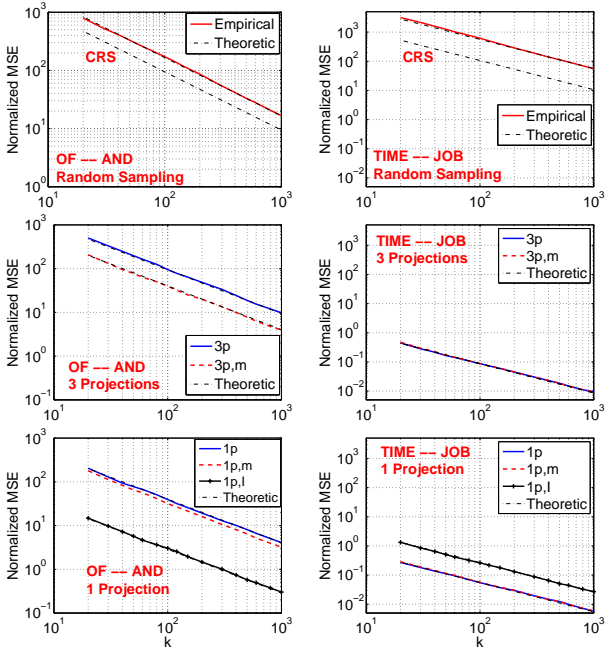

Figure 3: **Experiment 1**. *OF–AND* and *TIME–JOB*. See the caption of Figure 2 for explanations.

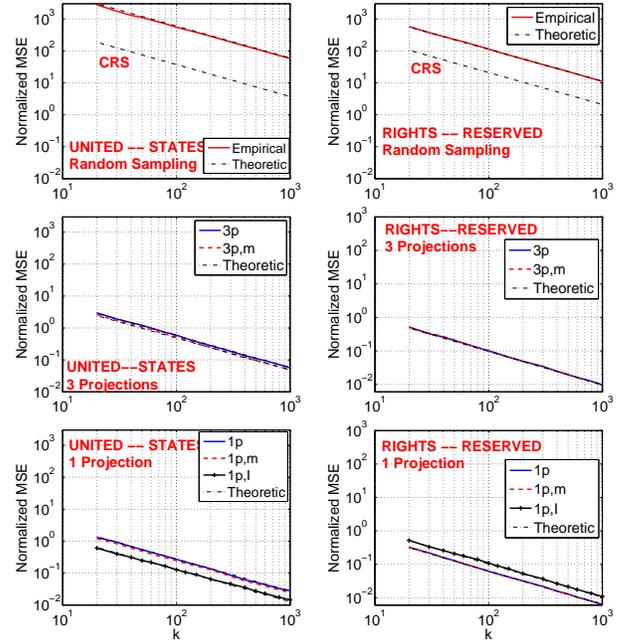

Figure 5: **Experiment 1**. *UNITED–STATES* and *RIGHTS–RESERVED*.

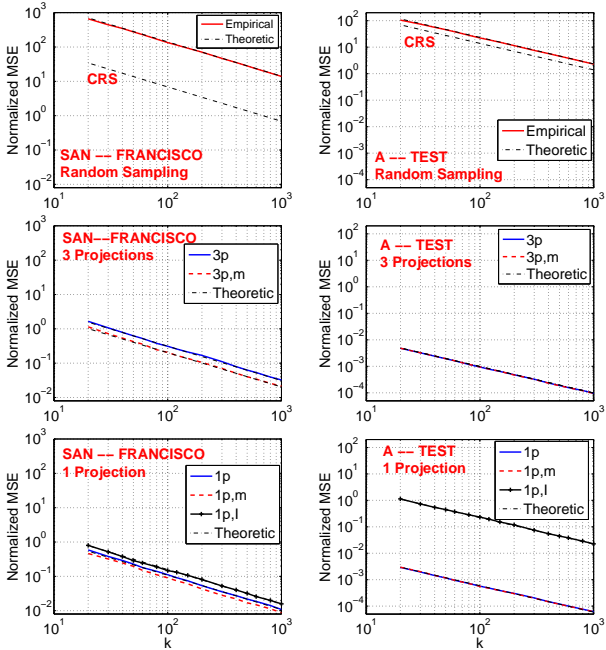

Figure 4: **Experiment 1**. *SAN–FRANCISCO* and *A–TEST*.

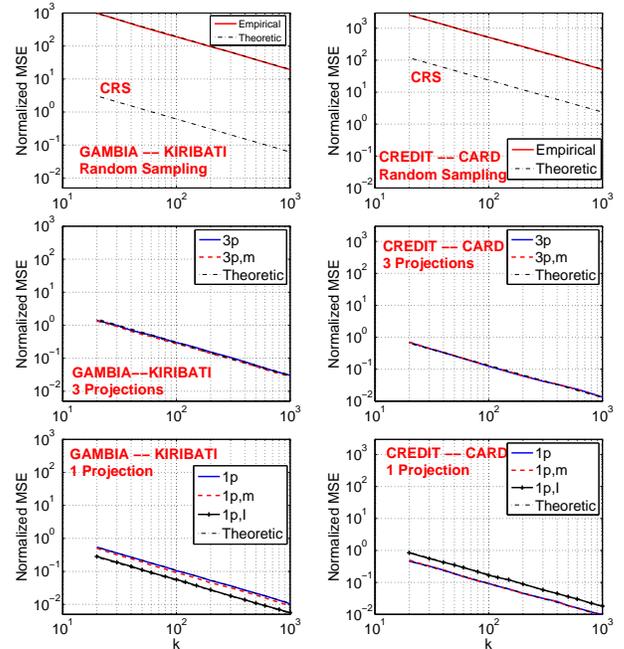

Figure 6: **Experiment 1**. *GAMBIA–KIRIBATI* and *CREDIT–CARD*.

### 4.3 Experiment 3

We also applied $m$-nearest neighbor method to classify the *Realsim* dataset, using the estimated $l_4$ distances. Again, only $\hat{d}_{(4),1p}$ and $\hat{d}_{(4),1p,m}$ were used.

Figure 8 reports the experiment results. Compared to **Experiment 2**, using margin on this dataset does not help reduce the classification errors. Interestingly, when $m = 10$ and $m = 20$ (bottom panels), we can see that using the estimated distances can actually result in smaller classification errors compared to using the original distances.

This interesting phenomenon, of course, may not be totally unexpected. There is really no strict relation that using more accurate distances will always produce smaller **test** classification errors. In fact, using the estimated distances may even have the beneficial "regularization" effect because it is in a sense less greedy.

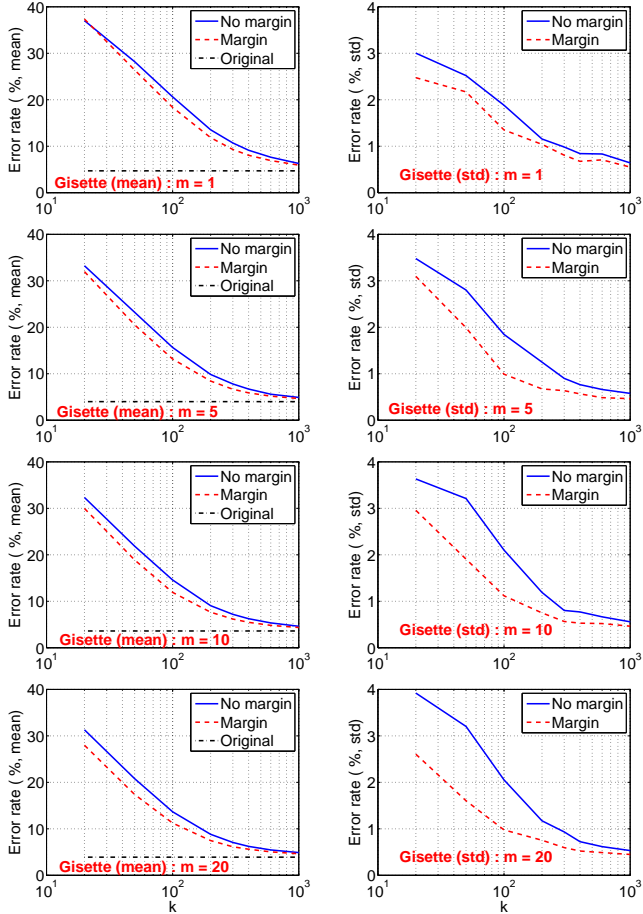

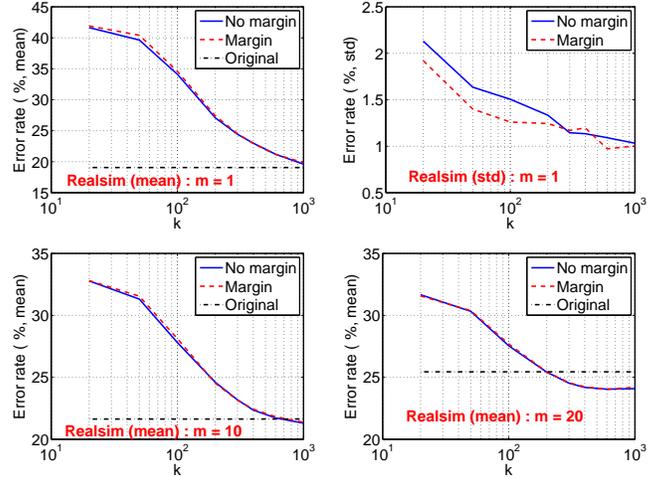

Figure 8: **Experiment 3**. We applied random projections (1 projection matrix) to estimate the $l_4$ distances and used them for $m$-nearest neighbor classification on the *Realsim* dataset. We only reported $m = 1$, $m = 10$, and $m = 20$.

Figure 7: **Experiment 2**. We applied random projections (1 projection matrix) to estimate the $l_4$ distances and used them for $m$-nearest neighbor classification on the *Gisette* dataset. We repeated the experiment 100 times and report the means (left panels) and standard errors (right panels) of the classification error rates, together with the classification errors using the original $l_4$ distances (dashed horizontal lines). We reported $m = 1$, $m = 5$, $m = 10$, and $m = 20$. In the legends, "No margin" denotes the estimator $\hat{d}_{(4),1p}$ and "margin" denotes the estimator $\hat{d}_{(4),1p,m}$.

## 5 Discussions

### 5.1 Extension to $p = 6, 8, 10, ...$

There is a straightforward extension to $p = 6, 8, 10, ...$. For example, when $p = 6$,

$$d_{(6)} = \sum_{i=1}^{D} |x_i - y_i|^6 = \sum_{i=1}^{D} x_i^6 + \sum_{i=1}^{D} y_i^6 - 20 \sum_{i=1}^{D} x_i^3 y_i^3$$
$$+ 15 \sum_{i=1}^{D} x_i^2 y_i^4 + 15 \sum_{i=1}^{D} x_i^4 y_i^2 - 6 \sum_{i=1}^{D} x_i^5 y_i - 6 \sum_{i=1}^{D} x_i y_i^5$$

We can again use normal random projections to estimate these five inner products: $\sum_{i=1}^{D} x_i^3 y_i^3$, $\sum_{i=1}^{D} x_i y_i^5$ etc.

There are many theoretical problems for future research. For example, should we use 5 projection matrices, or only 1 projection matrix, or something in between? How should we take advantage of the marginal information, etc.

On the other hand, we expect the $l_4$ distance will be the most useful in practice (after $l_2$). The kurtosis statistic is commonly used but the uses of higher (than 4th) order statistics are rare. From the approximation point of view, since the (worst case) complexity lower bound is $\Omega\left(D^{1-2/p}\right)$, it requires significantly more samples to approximate distances when $p > 4$.

### 5.2 Our Algorithms Attain the Lower Bound

The worst case complexity lower bound for approximating the $l_p$ norm was proved [2, 24] to be $k = \Omega\left(D^{1-2/p}\right)$, when $p > 2$. Here, we use the standard technique (which is popular in the Theory literature) to argue that our algorithms attain this worst case lower bound.

Consider a random variable $z$. Suppose $E(z) = \mu$ and $Var(z)$ is finite. In order to estimate $\mu$ with a guaranteed accuracy, it suffices to use $k = O\left(\frac{Var(z)}{\mu^2}\right)$ samples (suppressing $\epsilon$ and other constants). [15] intensively used this technique to analyze the complexity bound of their algorithms for approximating the matrix permanent.

In our case, for $p = 4$, the variances, e.g., (6), are complicated. However, it suffices to study the dominating terms, i.e., $\sum_{i=1}^{D} x_i^2 \sum_{i=1}^{D} y_i^6$ and $\sum_{i=1}^{D} x_i^6 \sum_{i=1}^{D} y_i^2$.

The dominating terms of the $l_4$ distance are $\sum_{i=1}^{D} x_i^4$ and $\sum_{i=1}^{D} y_i^4$. Because we are only interested in the order, not the precise constant, it suffices to use the following ratio as

the sample complexity of our algorithm for estimating $d_{(4)}$:

$$Z_{(4)} = \frac{\sum_{i=1}^{D} z_i^2 \sum_{i=1}^{D} z_i^6}{\left(\sum_{i=1}^{D} z_i^4\right)^2} \quad (18)$$

We can argue that $Z_{(4)} = O\left(D^{1/2}\right)$, i.e., the lower bound $\Omega\left(D^{1-2/p}\right)$ is attained for $p = 4$. In fact, we can even argue directly for general $p = 4, 6, 8, ...$, not just $p = 4$.

For $p = 4, 6, 8, ...$, we need to study the growth rate of

$$Z_{(p)} = \frac{\sum_{i=1}^{D} z_i^2 \sum_{i=1}^{D} z_i^{2p-2}}{\left(\sum_{i=1}^{D} z_i^p\right)^2}. \quad (19)$$

According to Holder's Inequality, if $1/(p/2) + 1/q = 1$, then

$$\sum_{i=1}^{D} z_i^2 \leq \left[\sum_{i=1}^{D} z_i^{2p/2}\right]^{2/p} \times \left[\sum_{i=1}^{D} 1^q\right]^{1/q} = \left[\sum_{i=1}^{D} z_i^p\right]^{2/p} D^{1-2/p}$$

$$Z_{(p)} = \frac{\sum_{i=1}^{D} z_i^2 \sum_{i=1}^{D} z_i^{2p-2}}{\left(\sum_{i=1}^{D} z_i^p\right)^2} \leq \frac{\sum_{i=1}^{D} z_i^{2p-2}}{\left[\sum_{i=1}^{D} z_i^p\right]^{2-2/p}} D^{1-2/p}$$

$$\leq \frac{\sum_{i=1}^{D} z_i^{2p-2}}{\sum_{i=1}^{D} z_i^{p(2-2/p)}} D^{1-2/p} = D^{1-2/p}.$$

Therefore, our algorithms (for all $p = 4, 6, 8, ...$) are optimal from the computational complexity perspective.

### 5.3 Sampling from Normal-Like Distributions

In our algorithms, it is not essential to sample the projection matrices from normal distributions. In fact, any zero-mean normal-like (i.e., with finite variance) distributions will suffice, for example, the well-known 3-point distribution[1]: $\{-1, 0, 1\}$ with probabilities $\{1/6, 2/3, 1/6\}$, or even using a very sparse projection matrix [22].

According to the central limit theorem (because $D$ is very large), sampling from any zero-mean normal-like distributions will result in essentially the same performance as far as the (asymptotic) estimation accuracy is concerned.

## 6 Conclusion

We provide simple algorithms for efficiently computing (estimating) $l_p$ $(p > 2)$ distances in large data matrices. While we mainly focus on $l_4$, our methods extend naturally to $l_6, l_8, l_{10}, ...$, i.e., when $p$ is even. Closely related to the kurtosis, the $l_4$ distance is an important summary statistic. We experimented with our algorithms on various datasets.

Our contributions are both theoretical and practical. Efficiently approximating the $l_p$ distances with $p > 2$ is a known difficult task. Our proposed simple algorithms are optimal in the sense that they attain the theoretical (worst case) complexity lower bound. Our algorithms are also practical, because they are simple, accurate, and there are practical needs for approximating $l_p$ distances with $p > 2$.

## References

[1] Dimitris Achlioptas. Database-friendly random projections. In *PODS*, pages 274–281, 2001.

[2] Ziv Bar-Yossef, T. S. Jayram, Ravi Kumar, and D. Sivakumar. An information statistics approach to data stream and communication complexity. In *FOCS*, pages 209–218, 2002.

[3] Ella Bingham and Heikki Mannila. Random projection in dimensionality reduction: Applications to image and text data. In *KDD*, pages 245–250, 2001.

[4] Léon Bottou, Olivier Chapelle, Dennis DeCoste, and Jason Weston, editors. *Large-Scale Kernel Machines*. The MIT Press, 2007.

[5] Olivier Chapelle, Patrick Haffner, and Vladimir N. Vapnik. Support vector machines for histogram-based image classification. *IEEE Trans. Neural Networks*, 10(5):1055–1064, 1999.

[6] Sanjoy Dasgupta and Yoav Freund. Random projection trees and low dimensional manifolds. In *STOC*, pages 537–546, 2008.

[7] Sanjoy Dasgupta and Anupam Gupta. An elementary proof of a theorem of Johnson and Lindenstrauss. *Random Structures and Algorithms*, 22(1):60 – 65, 2003.

[8] Dmitriy Fradkin and David Madigan. Experiments with random projections for machine learning. In *KDD*, pages 517–522, 2003.

[9] Yoav Freund, Sanjoy Dasgupta, Mayank Kabra, and Nakul Verma. Learning the structure of manifolds using random projections. In *NIPS*, 2008.

[10] Jonathan Goldstein, John C. Platt, and Christopher J. C. Burges. Redundant bit vectors for quickly searching high-dimensional regions. In *Deterministic and Statistical Methods in Machine Learning*, pages 137–158, 2004.

[11] Aapo Hyvrinen, Juha Karhunen, and Erkki Oja. *Independent Component Analysis*. John Wiley & Sons, 2001.

[12] Piotr Indyk. Stable distributions, pseudorandom generators, embeddings, and data stream computation. *Journal of ACM*, 53(3):307–323, 2006.

[13] Piotr Indyk and David P. Woodruff. Optimal approximations of the frequency moments of data streams. In *STOC*, pages 202–208, 2005.

[14] William B. Johnson and Joram Lindenstrauss. Extensions of Lipschitz mapping into Hilbert space. *Contemporary Mathematics*, 26:189–206, 1984.

[15] N. Karmarkar, R. Karp, R. Lipton, L. Lovasz, and M. Luby. A monte-carlo algorithm for estimating the permanent. *SIAM J. Comput.*, 22(2):284–293, 1993.

[16] Edwin M. Knorr, Raymond T. Ng, and Vladimir Tucakov. Distance-based outliers: algorithms and applications. *VLDB Journal*, 8(3-4):237–253, 2000.

[17] Ping Li. Computationally efficient estimators for dimension reductions using stable random projections. In *ICDM*, 2008.

[18] Ping Li. Estimators and tail bounds for dimension reduction in $l_\alpha$ ($0 < \alpha \leq 2$) using stable random projections. In *SODA*, pages 10 – 19, 2008.

[19] Ping Li and Kenneth W. Church. A sketch algorithm for estimating two-way and multi-way associations. *Computational Linguistics*, 33(3):305–354, 2007 (Preliminary results appeared in HLT/EMNLP 2005).

[20] Ping Li, Kenneth W. Church, and Trevor J. Hastie. One sketch for all: Theory and applications of conditional random sampling. In *NIPS*, 2008.

[21] Ping Li, Trevor J. Hastie, and Kenneth W. Church. Improving random projections using marginal information. In *COLT*, pages 635–649, 2006.

[22] Ping Li, Trevor J. Hastie, and Kenneth W. Church. Very sparse random projections. In *KDD*, pages 287–296, 2006.

[23] Ping Li and Arnd Christian König. b-bit minwise hashing. In *WWW*, pages 671–680, 2010.

[24] Michael E. Saks and Xiaodong Sun. Space lower bounds for distance approximation in the data stream model. In *STOC*, pages 360–369, 2002.

[25] Bernhard Schölkopf and Alexander J. Smola. *Learning with Kernels*. The MIT Press, 2002.

[26] Santosh Vempala. *The Random Projection Method*. American Mathematical Society, 2004.

# A  Proof of Lemma 6

We first decompose $\hat{d}_{(4),1p,I} = T + W$, where

$$T = \frac{1}{k}\sum_{j=1}^{k}\left\{-3u_{2,j}^2 - 3v_{2,j}^2 + 4u_{1,j}u_{3,j} + v_{1,j}v_{3,j}\right\}$$

$$W = \frac{1}{k}\sum_{j=1}^{k}\left\{6u_{2,j}v_{2,j} - 4u_{3,j}v_{1,j} - 4u_{1,j}v_{3,j}\right\}$$

Note that $\mathrm{Var}(W) = \mathrm{Var}\left(\hat{d}_{(4),1p}\right)$. To prove the unbiasedness, we need the expectations of the first-order terms.

$$u_{2,j}v_{2,j} = \sum_{i=1}^{D} x_i^2 r_{ij} \sum_{i=1}^{D} y_i^2 r_{ij} = \sum_{i=1}^{D} x_i^2 y_i^2 r_{ij}^2 + \sum_{i\neq i'} x_i^2 r_{ij} y_{i'}^2 r_{i'j}$$

$$\mathrm{E}(u_{2,j}v_{2,j}) = \sum_{i=1}^{D} \mathrm{E}\left(x_i^2 y_i^2 r_{ij}^2\right) + \sum_{i\neq i'} \mathrm{E}\left(x_i^2 r_{ij} y_{i'}^2 r_{i'j}\right) = \sum_{i=1}^{D} x_i^2 y_i^2,$$

$$\mathrm{E}(u_{3,j}v_{1,j}) = \sum_{i=1}^{D} x_i^3 y_i, \qquad \mathrm{E}(u_{1,j}v_{3,j}) = \sum_{i=1}^{D} x_i y_i^3,$$

$$\mathrm{E}\left(u_{2,j}^2\right) = \mathrm{E}(u_{1,j}u_{3,j}) = \sum_{i=1}^{D} x_i^4, \qquad \mathrm{E}\left(v_{2,j}^2\right) = \mathrm{E}(v_{1,j}v_{3,j}) = \sum_{i=1}^{D} y_i^4,$$

which allow us to prove that $\hat{d}_{(4),1p,I}$ is unbiased:

$$\mathrm{E}\left(\hat{d}_{(4),1p,I}\right) = -3\sum_{i=1}^{D}x_i^4 - 3\sum_{i=1}^{D}y_i^4 + 4\sum_{i=1}^{D}x_i^4 + 4\sum_{i=1}^{D}y_i^4$$
$$+ 6\sum_{i=1}^{D} x_i^2 y_i^2 - 4\sum_{i=1}^{D} x_i^3 y_i - 4\sum_{i=1}^{D} x_i y_i^3 = d_{(4)}.$$

To derive the variance, we need the expectations of the 2nd order terms. We first derive a useful general formula:

$$\mathrm{E}\left(\sum_{i=1}^{D} a_i r_{ij} \sum_{i=1}^{D} b_i r_{ij} \sum_{i=1}^{D} c_i r_{ij} \sum_{i=1}^{D} d_i r_{ij}\right) \qquad (20)$$

$$=\mathrm{E}\left(\left(\sum_{i=1}^{D} a_i b_i r_{ij}^2 + \sum_{i\neq i'} a_i r_{ij} b_{i'} r_{i'j}\right)\left(\sum_{i=1}^{D} c_i d_i r_{ij}^2 + \sum_{i\neq i'} c_i r_{ij} d_{i'} r_{i'j}\right)\right)$$

$$=\mathrm{E}\left(\sum_{i=1}^{D} a_i b_i c_i d_i r_{ij}^4 + \sum_{i\neq i'} a_i b_i r_{ij}^2 c_{i'} d_{i'} r_{i'j}^2\right)$$

$$+\mathrm{E}\left(\sum_{i\neq i'} a_i c_i r_{ij}^2 b_{i'} d_{i'} r_{i'j}^2 + \sum_{i\neq i'} a_i d_i r_{ij}^2 b_{i'} c_{i'} r_{i'j}^2\right)$$

$$=3\sum_{i=1}^{D} a_i b_i c_i d_i + \sum_{i\neq i'} a_i b_i c_{i'} d_{i'} + \sum_{i\neq i'} a_i c_i b_{i'} d_{i'} + \sum_{i\neq i'} a_i d_i b_{i'} c_{i'}$$

$$=\sum_{i} a_i b_i \sum_{i} c_i d_i + \sum_{i} a_i c_i \sum_{i=1}^{D} b_i d_i + \sum_{i} a_i d_i \sum_{i=1}^{D} b_i c_i.$$

We now focus on deriving the variance of $W$.

$$(6u_{2,j}v_{2,j} - 4u_{3,j}v_{1,j} - 4u_{1,j}v_{3,j})^2$$
$$=36u_{2,j}^2 v_{2,j}^2 + 16u_{3,j}^2 v_{1,j}^2 + 16u_{1,j}^2 v_{3,j}^2 - 48u_{2,j}u_{3,j}v_{2,j}v_{1,j}$$
$$- 48u_{2,j}u_{1,j}v_{2,j}v_{3,j} + 32u_{3,j}u_{1,j}v_{1,j}v_{3,j}.$$

From the generic formula (20), we know

$$\mathrm{E}\left(u_{2,j}^2 v_{2,j}^2\right) = \sum_{i=1}^{D} x_i^4 \sum_{i=1}^{D} y_i^4 + 2\left(\sum_{i=1}^{D} x_i^2 y_i^2\right)^2,$$

$$\mathrm{E}\left(u_{3,j}^2 v_{1,j}^2\right) = \sum_{i=1}^{D} x_i^6 \sum_{i=1}^{D} y_i^2 + 2\left(\sum_{i=1}^{D} x_i^3 y_i\right)^2,$$

$$\mathrm{E}\left(u_{1,j}^2 v_{3,j}^2\right) = \sum_{i=1}^{D} x_i^2 \sum_{i=1}^{D} y_i^6 + 2\left(\sum_{i=1}^{D} x_i y_i^3\right)^2,$$

$$\mathrm{E}(u_{2,j}u_{3,j}v_{2,j}v_{1,j})$$
$$=\sum_{i=1}^{D} x_i^5 \sum_{i=1}^{D} y_i^3 + \sum_{i=1}^{D} x_i^2 y_i^2 \sum_{i=1}^{D} x_i^3 y_i + \sum_{i=1}^{D} x_i^2 y_i \sum_{i=1}^{D} x_i^3 y_i^2,$$

$$\mathrm{E}(u_{2,j}u_{1,j}v_{2,j}v_{3,j})$$
$$=\sum_{i=1}^{D} x_i^3 \sum_{i=1}^{D} y_i^5 + \sum_{i=1}^{D} x_i y_i^3 \sum_{i=1}^{D} x_i^2 y_i^2 + \sum_{i=1}^{D} x_i y_i^2 \sum_{i=1}^{D} x_i^2 y_i^3,$$

$$\mathrm{E}(u_{3,j}u_{1,j}v_{1,j}v_{3,j})$$
$$=\sum_{i=1}^{D} x_i^4 \sum_{i=1}^{D} y_i^4 + \sum_{i=1}^{D} x_i y_i^3 \sum_{i=1}^{D} x_i^3 y_i + \sum_{i=1}^{D} x_i y_i \sum_{i=1}^{D} x_i^3 y_i^3.$$

Therefore,

$$\mathrm{Var}(6u_{2,j}v_{2,j} - 4u_{3,j}v_{1,j} - 4u_{1,j}v_{3,j})$$
$$=36\sum_{i=1}^{D} x_i^4 \sum_{i=1}^{D} y_i^4 + 72\left(\sum_{i=1}^{D} x_i^2 y_i^2\right)^2 + 16\sum_{i=1}^{D} x_i^6 \sum_{i=1}^{D} y_i^2$$
$$+32\left(\sum_{i=1}^{D} x_i^3 y_i\right)^2 + 16\sum_{i=1}^{D} x_i^2 \sum_{i=1}^{D} y_i^6 + 32\left(\sum_{i=1}^{D} x_i y_i^3\right)^2$$
$$-48\left(\sum_{i=1}^{D} x_i^5 \sum_{i=1}^{D} y_i^3 + \sum_{i=1}^{D} x_i^2 y_i^2 \sum_{i=1}^{D} x_i^3 y_i + \sum_{i=1}^{D} x_i^2 y_i \sum_{i=1}^{D} x_i^3 y_i^2\right)$$
$$-48\left(\sum_{i=1}^{D} x_i^3 \sum_{i=1}^{D} y_i^5 + \sum_{i=1}^{D} x_i y_i^3 \sum_{i=1}^{D} x_i^2 y_i^2 + \sum_{i=1}^{D} x_i y_i^2 \sum_{i=1}^{D} x_i^2 y_i^3\right)$$
$$+32\left(\sum_{i=1}^{D} x_i^4 \sum_{i=1}^{D} y_i^4 + \sum_{i=1}^{D} x_i y_i^3 \sum_{i=1}^{D} x_i^3 y_i + \sum_{i=1}^{D} x_i y_i \sum_{i=1}^{D} x_i^3 y_i^3\right)$$
$$-\left(6\sum_{i=1}^{D} x_i^2 y_i^2 - 4\sum_{i=1}^{D} x_i^3 y_i - 4\sum_{i=1}^{D} x_i y_i^3\right)^2$$

$$\mathrm{Var}(W) = \mathrm{Var}\left(\hat{d}_{(4),1p}\right)$$
$$=\frac{36}{k}\left(\sum_{i=1}^{D} x_i^4 \sum_{i=1}^{D} y_i^4 + \left(\sum_{i=1}^{D} x_i^2 y_i^2\right)^2\right)$$
$$+\frac{16}{k}\left(\sum_{i=1}^{D} x_i^6 \sum_{i=1}^{D} y_i^2 + \left(\sum_{i=1}^{D} x_i^3 y_i\right)^2\right)$$
$$+\frac{16}{k}\left(\sum_{i=1}^{D} x_i^2 \sum_{i=1}^{D} y_i^6 + \left(\sum_{i=1}^{D} x_i y_i^3\right)^2\right)$$
$$-\frac{48}{k}\left(\sum_{i=1}^{D} x_i^5 \sum_{i=1}^{D} y_i^3 + \sum_{i=1}^{D} x_i^2 y_i \sum_{i=1}^{D} x_i^3 y_i^2\right)$$
$$-\frac{48}{k}\left(\sum_{i=1}^{D} x_i^3 \sum_{i=1}^{D} y_i^5 + \sum_{i=1}^{D} x_i y_i^2 \sum_{i=1}^{D} x_i^2 y_i^3\right)$$
$$+\frac{32}{k}\left(\sum_{i=1}^{D} x_i^4 \sum_{i=1}^{D} y_i^4 + \sum_{i=1}^{D} x_i y_i \sum_{i=1}^{D} x_i^3 y_i^3\right)$$

We now work on the variance of $\mathrm{Var}(T)$.

$$(-3u_{2,j}^2 - 3v_{2,j}^2 + 4u_{1,j}u_{3,j} + 4v_{1,j}v_{3,j})^2$$
$$=9u_{2,j}^4 + 9v_{2,j}^4 + 16u_{1,j}^2 u_{3,j}^2 + 16v_{1,j}^2 v_{3,j}^2 + 18u_{2,j}^2 v_{2,j}^2 - 24u_{2,j}^2 u_{1,j}u_{3,j}$$
$$-24u_{2,j}^2 v_{1,j}v_{3,j} - 24v_{2,j}^2 u_{1,j}u_{3,j} - 24v_{2,j}^2 v_{1,j}v_{3,j} + 32u_{1,j}u_{3,j}v_{1,j}v_{3,j}$$

Again, from the generic formula (20), we obtain

$$\mathrm{E}\left(u_{2,j}^4\right) = 3\left(\sum_{i=1}^{D} x_i^4\right)^2, \quad \mathrm{E}\left(v_{2,j}^4\right) = 3\left(\sum_{i=1}^{D} y_i^4\right)^2,$$

$$\mathrm{E}\left(v_{1,j}^2 v_{3,j}^2\right) = \sum_{i=1}^{D} y_i^2 \sum_{i=1}^{D} y_i^6 + 2\left(\sum_{i=1}^{D} y_i^4\right)^2,$$

$$\mathrm{E}\left(u_{1,j}^2 u_{3,j}^2\right) = \sum_{i=1}^{D} x_i^2 \sum_{i=1}^{D} x_i^6 + 2\left(\sum_{i=1}^{D} x_i^4\right)^2,$$

$$\mathrm{E}\left(u_{2,j}^2 u_{1,j}u_{3,j}\right) = \left(\sum_{i=1}^{D} x_i^4\right)^2 + 2\sum_{i=1}^{D} x_i^3 \sum_{i=1}^{D} x_i^5,$$

$$\mathrm{E}\left(v_{2,j}^2 v_{1,j}v_{3,j}\right) = \left(\sum_{i=1}^{D} y_i^4\right)^2 + 2\sum_{i=1}^{D} y_i^3 \sum_{i=1}^{D} y_i^5,$$

$$\mathrm{E}\left(u_{2,j}^2 v_{1,j}v_{3,j}\right) = \sum_{i=1}^{D} x_i^4 \sum_{i=1}^{D} y_i^4 + 2\sum_{i=1}^{D} x_i^2 y_i \sum_{i=1}^{D} x_i^2 y_i^3$$

$$\mathrm{E}\left(v_{2,j}^2 u_{1,j} u_{3,j}\right) = \sum_{i=1}^D x_i^4 \sum_{i=1}^D y_i^4 + 2\sum_{i=1}^D x_i y_i^2 \sum_{i=1}^D x_i^3 y_i^2$$

$$\mathrm{E}\left(u_{1,j} u_{3,j} v_{1,j} v_{3,j}\right)$$
$$= \sum_{i=1}^D x_i^4 \sum_{i=1}^D y_i^4 + \sum_{i=1}^D x_i y_i \sum_{i=1}^D x_i^3 y_i^3 + \sum_{i=1}^D x_i y_i^3 \sum_{i=1}^D x_i^3 y_i$$

Therefore,

$$\mathrm{Var}\left(-3u_{2,j}^2 - 3v_{2,j}^2 + 4u_{1,j}u_{3,j} + 4v_{1,j}v_{3,j}\right)$$
$$= 27\left(\sum_{i=1}^D x_i^4\right)^2 + 27\left(\sum_{i=1}^D y_i^4\right)^2 + 16\sum_{i=1}^D x_i^2 \sum_{i=1}^D x_i^6 + 32\left(\sum_{i=1}^D x_i^4\right)^2$$
$$+16\sum_{i=1}^D y_i^2 \sum_{i=1}^D y_i^6 + 32\left(\sum_{i=1}^D y_i^4\right)^2 + 18\sum_{i=1}^D x_i^4 \sum_{i=1}^D y_i^4 + 36\left(\sum_{i=1}^D x_i^2 y_i^2\right)^2$$
$$-24\left(\sum_{i=1}^D x_i^4\right)^2 - 48\sum_{i=1}^D x_i^3 \sum_{i=1}^D x_i^5 - 24\sum_{i=1}^D x_i^4 \sum_{i=1}^D y_i^4 - 48\sum_{i=1}^D x_i^2 y_i \sum_{i=1}^D x_i^2 y_i^3$$
$$-24\left(\sum_{i=1}^D y_i^4\right)^2 - 48\sum_{i=1}^D y_i^3 \sum_{i=1}^D y_i^5 - 24\sum_{i=1}^D x_i^4 \sum_{i=1}^D y_i^4 - 48\sum_{i=1}^D x_i y_i^2 \sum_{i=1}^D x_i^3 y_i^2$$
$$+32\sum_{i=1}^D x_i^4 \sum_{i=1}^D y_i^4 + 32\sum_{i=1}^D x_i y_i \sum_{i=1}^D x_i^3 y_i^3 + 32\sum_{i=1}^D x_i y_i^3 \sum_{i=1}^D x_i^3 y_i$$
$$-\left(\sum_{i=1}^D x_i^4 + \sum_{i=1}^D y_i^4\right)^2$$
$$= 34\left(\sum_{i=1}^D x_i^4\right)^2 + 34\left(\sum_{i=1}^D y_i^4\right)^2 + 16\sum_{i=1}^D x_i^2 \sum_{i=1}^D x_i^6 + 16\sum_{i=1}^D y_i^2 \sum_{i=1}^D y_i^6$$
$$+36\left(\sum_{i=1}^D x_i^2 y_i^2\right)^2 - 48\sum_{i=1}^D x_i^3 \sum_{i=1}^D x_i^5 - 48\sum_{i=1}^D x_i^2 y_i \sum_{i=1}^D x_i^2 y_i^3$$
$$-48\sum_{i=1}^D y_i^3 \sum_{i=1}^D y_i^5 - 48\sum_{i=1}^D x_i y_i^2 \sum_{i=1}^D x_i^3 y_i^2$$
$$+32\sum_{i=1}^D x_i y_i \sum_{i=1}^D x_i^3 y_i^3 + 32\sum_{i=1}^D x_i y_i^3 \sum_{i=1}^D x_i^3 y_i$$

Combining all results yields

$$k \times \mathrm{Var}\left(\hat{d}_{(4),1p,I}\right) = k \times \mathrm{Var}\left(T+W\right) = k \times \mathrm{Var}\left(\hat{d}_{(4),1p}\right)$$
$$+34\left(\sum_{i=1}^D x_i^4\right)^2 + 34\left(\sum_{i=1}^D y_i^4\right)^2 + 36\left(\sum_{i=1}^D x_i^2 y_i^2\right)^2 + 32\sum_{i=1}^D x_i y_i^3 \sum_{i=1}^D x_i^3 y_i$$
$$-72\sum_{i=1}^D x_i^4 \sum_{i=1}^D x_i^2 y_i^2 - 72\sum_{i=1}^D y_i^4 \sum_{i=1}^D x_i^2 y_i^2$$
$$-32\sum_{i=1}^D x_i^4 \sum_{i=1}^D x_i^3 y_i - 32\sum_{i=1}^D y_i^4 \sum_{i=1}^D x_i^3 y_i$$
$$-32\sum_{i=1}^D x_i^4 \sum_{i=1}^D x_i y_i^3 - 32\sum_{i=1}^D y_i^4 \sum_{i=1}^D x_i y_i^3$$
$$-48\sum_{i=1}^D x_i^3 \sum_{i=1}^D x_i^5 - 48\sum_{i=1}^D x_i^2 y_i \sum_{i=1}^D x_i^2 y_i^3$$
$$-48\sum_{i=1}^D y_i^3 \sum_{i=1}^D y_i^5 - 48\sum_{i=1}^D x_i y_i^2 \sum_{i=1}^D x_i^3 y_i^2$$
$$+48\sum_{i=1}^D x_i^3 \sum_{i=1}^D x_i^3 y_i^2 + 48\sum_{i=1}^D x_i^5 \sum_{i=1}^D x_i y_i^2$$
$$+48\sum_{i=1}^D y_i^3 \sum_{i=1}^D x_i^2 y_i^3 + 48\sum_{i=1}^D y_i^5 \sum_{i=1}^D x_i^2 y_i$$
$$+48\sum_{i=1}^D x_i^5 \sum_{i=1}^D x_i^2 y_i + 48\sum_{i=1}^D y_i^3 \sum_{i=1}^D x_i^3 y_i^2$$
$$+48\sum_{i=1}^D x_i^3 \sum_{i=1}^D x_i^2 y_i^3 + 48\sum_{i=1}^D y_i^5 \sum_{i=1}^D x_i y_i^2$$
$$-32\sum_{i=1}^D x_i^6 \sum_{i=1}^D x_i y_i - 32\sum_{i=1}^D y_i^2 \sum_{i=1}^D x_i^3 y_i^3 + 32\sum_{i=1}^D x_i y_i \sum_{i=1}^D x_i^3 y_i^3$$
$$-32\sum_{i=1}^D x_i^2 \sum_{i=1}^D x_i^3 y_i^3 - 32\sum_{i=1}^D y_i^6 \sum_{i=1}^D x_i y_i$$
$$+16\sum_{i=1}^D x_i^2 \sum_{i=1}^D x_i^6 + 16\sum_{i=1}^D y_i^2 \sum_{i=1}^D y_i^6$$

Next, we study $\mathrm{Cov}(T,W)$.

$$\left(6u_{2,j}v_{2,j} - 4u_{3,j}v_{1,j} - 4u_{1,j}v_{3,j}\right)\left(-3u_{2,j}^2 - 3v_{2,j}^2 + 4u_{1,j}u_{3,j} + 4v_{1,j}v_{3,j}\right)$$
$$= -18u_{2,j}^3 v_{2,j} - 18u_{2,j}v_{2,j}^3 + 24u_{2,j}v_{2,j}u_{1,j}u_{3,j} + 24u_{2,j}v_{2,j}v_{1,j}v_{3,j}$$
$$+12u_{3,j}v_{1,j}u_{2,j}^2 + 12u_{3,j}v_{1,j}v_{2,j}^2 - 16u_{3,j}^2 v_{1,j}u_{1,j} - 16u_{3,j}v_{1,j}^2 v_{3,j}$$
$$+12u_{1,j}v_{3,j}u_{2,j}^2 + 12u_{1,j}v_{3,j}v_{2,j}^2 - 16u_{1,j}^2 v_{3,j}u_{3,j} - 16u_{1,j}v_{1,j}v_{3,j}^2$$

$$\mathrm{Cov}\left[\left(6u_{2,j}v_{2,j} - 4u_{3,j}v_{1,j} - 4u_{1,j}v_{3,j}\right),\left(-3u_{2,j}^2 - 3v_{2,j}^2 + 4u_{1,j}u_{3,j} + 4v_{1,j}v_{3,j}\right)\right]$$
$$= -36\sum_{i=1}^D x_i^4 \sum_{i=1}^D x_i^2 y_i^2 - 36\sum_{i=1}^D y_i^4 \sum_{i=1}^D x_i^2 y_i^2 + 24\sum_{i=1}^D x_i^3 \sum_{i=1}^D x_i^3 y_i^2 + 24\sum_{i=1}^D x_i^5 \sum_{i=1}^D x_i y_i^2$$
$$+24\sum_{i=1}^D y_i^3 \sum_{i=1}^D x_i^2 y_i^3 + 24\sum_{i=1}^D y_i^5 \sum_{i=1}^D x_i^2 y_i$$
$$-16\sum_{i=1}^D x_i^4 \sum_{i=1}^D x_i^3 y_i + 24\sum_{i=1}^D x_i^5 \sum_{i=1}^D x_i^2 y_i - 16\sum_{i=1}^D y_i^4 \sum_{i=1}^D x_i^3 y_i + 24\sum_{i=1}^D y_i^3 \sum_{i=1}^D x_i^3 y_i^2$$
$$-16\sum_{i=1}^D x_i^6 \sum_{i=1}^D x_i y_i - 16\sum_{i=1}^D y_i^2 \sum_{i=1}^D x_i^3 y_i^3$$
$$-16\sum_{i=1}^D x_i^4 \sum_{i=1}^D x_i y_i^3 + 24\sum_{i=1}^D x_i^3 \sum_{i=1}^D x_i^2 y_i^3 - 16\sum_{i=1}^D y_i^4 \sum_{i=1}^D x_i y_i^3 + 24\sum_{i=1}^D y_i^5 \sum_{i=1}^D x_i y_i^2$$
$$-16\sum_{i=1}^D x_i^2 \sum_{i=1}^D x_i^3 y_i^3 - 16\sum_{i=1}^D y_i^6 \sum_{i=1}^D x_i y_i$$